\documentclass[twoside,11pt, preprint]{article}

\usepackage{blindtext}

\usepackage{dmlr2e}

\newcommand\blfootnote[1]{%
  \begingroup
  \renewcommand\thefootnote{}\footnote{#1}%
  \addtocounter{footnote}{-1}%
  \endgroup
}

\usepackage{lastpage}
\dmlrheading{2}{2024}{1-\pageref{LastPage}}{2/2/2024; Revised N/A}{N/A}{21-0000}{Chanjun Park and Minsoo Khang and Dahyun Kim} 

\def\openreview{} 

\firstpageno{1}

\begin{document}

\title{Model-Based Data-Centric AI: Bridging the Divide Between Academic Ideals and Industrial Pragmatism}

\author{Chanjun Park$^{\dagger*}$\quad Minsoo Khang$^{\dagger}$,\quad    Dahyun Kim$^{\dagger}$  \\
Upstage \\
{\texttt{\{chanjun.park, mkhang, kdahyun\}@upstage.ai}}}


\maketitle

\begin{abstract}
\blfootnote{$^*$ Corresponding Authors \\
$^{\dagger}$ Equal Contribution}
This paper delves into the contrasting roles of data within academic and industrial spheres, highlighting the divergence between Data-Centric AI and Model-Agnostic AI approaches. We argue that while Data-Centric AI focuses on the primacy of high-quality data for model performance, Model-Agnostic AI prioritizes algorithmic flexibility, often at the expense of data quality considerations. This distinction reveals that academic standards for data quality frequently do not meet the rigorous demands of industrial applications, leading to potential pitfalls in deploying academic models in real-world settings. Through a comprehensive analysis, we address these disparities, presenting both the challenges they pose and strategies for bridging the gap. Furthermore, we propose a novel paradigm: Model-Based Data-Centric AI, which aims to reconcile these differences by integrating model considerations into data optimization processes. This approach underscores the necessity for evolving data requirements that are sensitive to the nuances of both academic research and industrial deployment. By exploring these discrepancies, we aim to foster a more nuanced understanding of data's role in AI development and encourage a convergence of academic and industrial standards to enhance AI's real-world applicability.
\end{abstract}


\section{Introduction}
The divergence in the conceptualization of `good data' between academic research and industrial application is not just prevalent but also significant, especially when considering the development and deployment of models across different domains. Academia tends to prioritize datasets that encapsulate the essence of a given task, aiming for a broad applicability that facilitates fair comparisons across different models. This approach, while valuable for theoretical exploration, often overlooks the specificities that datasets require to optimize models for specialized tasks. 

In contrast, industry focuses on curating data that directly improves model performance for particular applications, integrating complex, real-world nuances including edge cases and long-tail distributions~\citep{cha2023unlocking,choi2023dmops}. This discrepancy highlights that benchmark datasets, while foundational, may not suffice for the nuanced needs of practical applications~\citep{srivastava2022beyond}, necessitating model-informed data curation to truly enhance performance~\citep{gupta2021data}.

Our critique extends beyond the current scope of academic data curation, which often fails to account for the dynamic interaction between data and models. This oversight is not due to a lack of research but rather to the dominant focus on data-centric AI that neglects the intricate ways in which models engage with and learn from data~\citep{zha2023data}.

From an academic lens, creating datasets without considering specific model requirements might seem logical. However, for models in the industrial sphere, this practice is counterproductive. Datasets predominantly composed of straightforward positive examples do little to challenge or improve models, leading to marginal gains after reaching a certain threshold of data quantity ~\citep{soviany2022curriculum}. The complexity and depth of models necessitate datasets that not only cover a wide range of scenarios but also include challenging, model-specific instances to drive significant performance improvements.

Therefore, we propose a pivotal shift away from \textit{model-agnostic} data curation towards a \textit{Model-Based Data-Centric AI}. This paradigm shift is essential for aligning academic research with industrial objectives, ensuring that datasets are not only theoretically comprehensive but also practically effective in enhancing model performance. By integrating model-specific insights into the data curation process, we can bridge the existing gap, enabling models to achieve their full potential in both research and real-world applications.

Moreover, it is important to note that these considerations are equally applicable to the realm of Large Language Models (LLMs)~\citep{zhao2023survey}. As evidenced by works such as~\citep{madaan2023selfrefine}, the nuanced data requirements and model-specific considerations discussed herein are not confined to traditional AI models but extend to LLMs as well. Given their vast capabilities and varied applications, LLMs stand to benefit significantly from data curation practices that are informed by both the complexity of real-world data and the specific demands of the tasks they are designed to perform. This inclusive approach highlights the universal importance of sophisticated data curation strategies, underscoring the need for continuous evolution in our approach to data, to meet the demands of the ever-advancing AI landscape, including the burgeoning field of LLMs.

\section{Data-Centric AI and Model-Specific}
In the field of academia, datasets are often curated in a model-agnostic manner, to facilitate fair evaluation over various models. This paper argues that model-dependent features need to be considered in the curation process as well to produce the best performing \textit{industrial} models instead. Two main features to be discussed in this paper are: approach-specific information and model guided data synthesis.

\paragraph{Approach-specific information}. In curating datasets, annotating additional information that is not inherently fundamental to the task at hand can be beneficial for certain approaches. These additional annotations can range from metadata to additional labels depending of which can all benefit the downstream task solution despite not being essential to the target task.

For instance, let us discuss \textit{Document Visual Question Answering} (DocVQA) task (where document images, text transcriptions, and question-and-answer pairs are given) as an example~\citep{mathew2021docvqa}, where two broad groups of solutions exist: \textit{extractive} and \textit{generative}. Generative approaches aim to generate the answer for the given question conditioned on the input image. On the other hand, extractive methods assume that the answer to the given question is a substring of the text transcription and aim to extract or \textit{index} the answer from the given text transcription conditioned on the input image. Despite the high output capacity of generative approaches, extractive approaches are often preferred in use-cases where the tolerance for \textit{hallucinated} answers is low.

For generative approaches, document images, text transcriptions, and question-and-answer pairs are sufficient in providing the model with the necessary supervisory signals. Extractive approaches, on the other hand, require both the start and end indices of answer substrings in the given text transcriptions. Without these indices, extractive models cannot receive the supervisory signals for training, and the absence of such approach-specific information results in the need to re-label the dataset, that is not only costly but potentially impractical. Provisioning of such approach-specific information, beyond just task-specific ones, need to be thoroughly considered during the dataset curation stage, especially in the context of \textit{industrial} applications.

\paragraph{Model guided data synthesis}. Synthetic data generation is often used alongside real data curation to supplement the data deficiency and class imbalance issues~\citep{johnson2019survey,narayanan2022curator}. While these techniques behind data synthesis are helpful in general, it is often difficult to \textit{evaluate} the quality of the synthetic data aside from hand-picked qualitative comparisons. We argue that the impact of the synthetic data on the downstream model performance should be used as guidance in gaining insight on the quality of the synthetic data and potential improvements.

Taking character recognition task (where word images and corresponding word texts are given) as an example, character recognition in agglutinative languages such as Korean is often challenged with handling characters with long-tailed distribution~\citep{park2023improving}. A simple data synthesis method would be to synthesize more words using the tail part of the character distribution, \textit{i.e.,} up-sampling the tail characters. While such methods do benefit the model, the generated words are often meaningless and harms the language modeling performance of the trained character recognition models. Thus, the authors~\citep{park2023improving} showed that models trained on the naively generated dataset lacked word-level contextual awareness. The proposed synthetic data generation mixes the data created by up-sampling the tail characters with the original data created from a long-tail character distribution which benefited the model's overall performance. While this exemplifies how using downstream model performance as guidance for data synthesis processes is helpful, it still leaves a lot to be desired as the proposed solution leaves much room for improvement.

Recent works on fine-tuning large language models also show the potential of model guided synthesis in the form of self-refinement~\citep{madaan2023selfrefine}.
In essence, the model that needs to be improved is directly involved in the data generation process, which distinctly sets self-refinement from more academic methods of creating new data.
Additionally, as fine-tuning of LLMs expands towards to human preference alignment, various methods have been proposed to create new preference data~\citep{cui2023ultrafeedback, kim2023solar, yuan2024self}. The generated data all have strong dependencies with the specific model that was used in the synthesis process, yet harbors widespread usage owing to the practical effectiveness of the created data.
Thus, model guided data synthesis is becoming the norm in the rapidly evolving world of AI.

\section{Disparities in Data Quality Standards: Academic Rigor vs. Industrial Pragmatism}
The conceptual divergence between academia and industry regarding ``good data'' is both nuanced and significant, reflecting deeper methodological and practical differences in how data is curated, maintained, and evolved within these sectors. In academic circles, datasets are often treated as fixed benchmarks. Once a dataset is curated and publicly released, it typically remains unchanged for extended periods. Subsequent versions or enhancements are infrequent, reflecting a culture where the creation of a dataset is seen as a singular scholarly contribution. This approach can lead to notable delays in updates, with some of the most critical datasets in fields like computer vision waiting over a decade for a new iteration, often contingent upon the initiative of a completely different research group~\citep{kim2023inter,kim2023transcending}. This scenario underscores a fundamental academic assumption: that the primary value of a dataset lies in its initial creation and its ability to benchmark model performance at a point in time~\citep{recht2019imagenet,deng2009imagenet}.

Contrastingly, the industry adopts a dynamic and iterative perspective on data curation. Datasets are not static but are continuously enhanced through numerous versions, each iteration subjected to rigorous cleansing and modification processes. This relentless pursuit of data quality ensures that datasets remain relevant and reflective of evolving real-world complexities. In this context, good data transcends the traditional academic metrics of task relevance and dataset size. It also encompasses the richness of metadata, which meticulously documents each iteration's changes, providing transparency and traceability essential for ongoing model development and deployment.

Moreover, the industrial emphasis on the iterative cleansing of data introduces additional criteria for evaluating data quality. Factors such as the cost-effectiveness of the cleansing process, the scalability and automation of these procedures, and the generation of comprehensive metadata become pivotal. These criteria highlight the industrial commitment to efficiency, reproducibility, and operational excellence in data management.

The expansion of the notion of good data within the industrial setting to include these operational factors marks a significant departure from academic conventions. While academia primarily focuses on the theoretical and experimental utility of datasets, industry prioritizes practicality, operational efficiency, and adaptability to technological advancements and market demands. This expanded definition encompasses not only the intrinsic qualities of the data but also the processes by which data is curated, maintained, and utilized over time.

This broader, more dynamic conception of good data in industry compared to the relatively static view in academia contributes to the disparity in data quality definitions between these sectors. It highlights a critical need for dialogue and convergence between academic and industrial approaches to data curation, aiming to bridge the gap and foster datasets that are not only scientifically robust but also practically valuable for real-world applications. Such a collaborative approach could enhance the relevance, utility, and impact of data across both research and industry, paving the way for innovations that are deeply informed by both theoretical insight and practical exigencies.

\section{Conclusion}
The rapid evolution of AI applications within the industrial sector necessitates a reevaluation of our data management strategies. While academic perspectives on data-centric AI offer valuable insights, we propose that the development of industrial AI requires a synthesis of data-centric and model-specific approaches, underpinned by a redefined standard for data quality. This hybrid approach not only leverages the strengths of data-centric methodologies but also integrates model-specific features and approach-specific information, paving the way for innovative methods and enhanced data generation efficacy through model-guided synthesis. Furthermore, an essential component of this industrial paradigm is the development and maintenance of comprehensive metadata that documents each iteration of the data cleansing process. This transparency, coupled with a focus on the cost-effectiveness and automation of data curation, is paramount for setting a new industrial standard for what constitutes good data. In conclusion, the advancement of AI in industrial contexts demands a harmonization of data-centric principles with model-specific insights. By adopting standards that reflect both the importance of data quality and the nuances of model interaction, we can foster the growth of a truly data-centric AI ecosystem, optimized for rapid and efficient industrial development. This shift towards a more nuanced, integrated approach will be critical for realizing the full potential of AI applications in industry.

\impact{This research elucidates the divergent paradigms of ``Data-Centric AI'' and ``Model-Agnostic AI'' across academic and industrial landscapes, spotlighting the criticality of data quality standards and their implications for real-world AI applications. By dissecting the nuances between academic and industry data standards, our work sheds light on the existing gaps and underscores the necessity for a harmonized approach towards data quality, paving the way for ``Model-Based Data-Centric AI''.}



\vskip 0.2in
\bibliography{sample}

\end{document}